\documentclass[twocolumn, switch]{article}

\usepackage{arxiv}

\usepackage[utf8]{inputenc} 
\usepackage[T1]{fontenc}    
\usepackage{hyperref}       
\usepackage{url}            
\usepackage{booktabs}       
\usepackage{amsfonts}       
\usepackage{nicefrac}       
\usepackage{microtype}      
\usepackage[binary-units=true]{siunitx}
\usepackage{comment}
\usepackage{stfloats}
\usepackage{graphicx}
\usepackage{multirow}
\usepackage{makecell}
\usepackage{epsfig}
\usepackage{pifont}
\usepackage{booktabs} 
\usepackage{footnote}
\usepackage[english]{babel}
\usepackage[linesnumbered,algoruled]{algorithm2e}
\usepackage{bm}
\usepackage{multicol}
\usepackage{setspace}
\usepackage{booktabs}
\usepackage{setspace}
\usepackage{epsfig}
\usepackage{verbatim}
\usepackage{amsmath}
\usepackage{float}
\usepackage{subfigure}
\usepackage{amssymb}
\usepackage[english]{babel}
\usepackage{array}
\usepackage{color}
\usepackage[font=small,labelfont=bf]{caption}
\usepackage{url}
\usepackage{hhline}
\usepackage[binary-units=true]{siunitx}
\usepackage[linesnumbered,algoruled]{algorithm2e}
\usepackage[table]{xcolor}
\DeclareBinaryPrefix\kilo{K}{2}
\usepackage{enumitem}

\newcommand\copyrighttext{%
  \footnotesize \textbf{IEEE Copyright Notice.}
\textcopyright 2019 IEEE. Personal use of this material is permitted. Permission from IEEE must be obtained for all other uses, in any current or future media, including reprinting/republishing this material for advertising or promotional purposes, creating new collective works, for resale or redistribution to servers or lists, or reuse of any copyrighted component of this work in other works.}

\title{EAST: Encoding-Aware Sparse Training for\\ Deep Memory Compression of ConvNets\vspace{-0mm}}
\date{\vspace{-5ex}}

\author{
Matteo Grimaldi, Valentino Peluso, Andrea Calimera\\
Politecnico di Torino, 10129 Torino, Italy\vspace*{0mm}
}

\begin{document}
\twocolumn[ 
  \begin{@twocolumnfalse} 

\maketitle
\vspace{-1cm}

\begin{abstract}
The implementation of Deep Convolutional Neural Networks (ConvNets) on tiny end-nodes with limited non-volatile memory space calls for smart compression strategies capable of shrinking the footprint yet preserving predictive accuracy. There exist a number of strategies for this purpose, from those that play with the topology of the model or the arithmetic precision, e.g. pruning and quantization, to those that operate a model agnostic compression, e.g. weight encoding. The tighter the memory constraint, the higher the probability that these techniques alone cannot meet the requirement, hence more awareness and cooperation across different optimizations become mandatory. This work addresses the issue by introducing EAST, {\em Encoding-Aware Sparse Training}, a novel memory-constrained training procedure that leads quantized ConvNets towards deep memory compression. EAST implements an adaptive group pruning designed to maximize the compression rate of the weight encoding scheme (the LZ4 algorithm in this work). If compared to existing methods, EAST meets the memory constraint with lower sparsity, hence ensuring higher accuracy. Results conducted on a state-of-the-art ConvNet (ResNet-9) deployed on a low-power microcontroller (ARM Cortex-M4) validate the proposal.
\end{abstract}
\vspace{0.1cm}

\vspace{0.35cm}


\textbf{This is a Preprint}. Accepted to be Published in Proceedings of \textcopyright 2020 IEEE International Conference on Artificial Intelligence Circuits and Systems, March 23-25 2020, Genova, Italy. 

\vspace{0.15cm} 
\copyrighttext
\vspace{0.35cm}

\section{Introduction \& Motivations}\label{sec:intro} 
\end{@twocolumnfalse}]
The deployment of Convolutional Neural Networks (ConvNets) on the edge of the Internet-of-Things (IoT) is a challenge as tiny sensor nodes are powered by low-cost microcontroller units (MCU) with limited memory and computational resources. The lack of memory space is of particular relevance here. Indeed, even the most optimized ConvNets may require tens of \si{\mega\byte}s of pre-trained parameters, while for off-the-shelf MCUs the non-volatile memories range from \SI{32}{\kilo\byte} to \SI{2}{\mega\byte}. Moreover, flash memories are also used to host other routines to drive the sensors and/or to pre-process the data, reducing, even more, the available space. Not less important, a ConvNet is trained to process a single specific task, while multi-sensor applications (e.g.~\cite{seidenari2017deep}) have to process different sources of information, thus needing multiple ConvNets on board.

Several recent works investigated aggressive optimization techniques for memory compression. Among them, quantization via fixed-point representation is now considered a mandatory step. The use of 8-bit integers is a common choice for general-purpose MCUs \cite{lai2018cmsis} as it reduces the memory footprint up to 4$\times$ w.r.t. 32-bit floating-point with no, or negligible, accuracy loss. However, quantization alone might be not enough to fit stringent memory requirements. Sparse training via \textit{weight pruning}~\cite{han2015learning}\cite{zhu2017prune} is an additional strategy that can improve the compression if combined with some encoding scheme and/or when quantization is jointly applied~\cite{han2016deep}\cite{tung2018clip}. Sparse trained and quantized ConvNets get easier to be compressed by lossless encoding schemes as their weight tensors have long chains of zeros or repeated values.

\begin{figure}[!t]
     \centering
     \epsfig{file=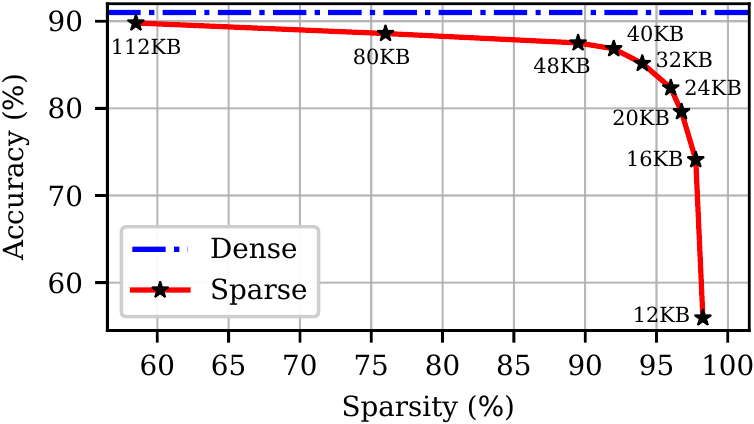,width=0.45\textwidth}
     \caption{Sparsity vs. Accuracy of a compressed 9-layer ResNet under different memory constraints (the labeled numbers). The net is trained on CIFAR-10, then compressed via weight pruning and encoding. The blue dash-dotted line marks the accuracy of the original dense version (\SI{140}{\kilo\byte}).}
     \label{fig:sparsity}
    \vspace*{-5mm}
\end{figure}

As a rule of thumb, the higher the sparsity, the larger the compression rate. However, under stringent constraints, i.e. a few tens of \si{\kilo\byte}s, the level of sparsity which will let encoding schemes meet the constraint is extremely high, much higher than what ConvNets may tolerate. Fig. \ref{fig:sparsity} illustrates such an important aspect for a 9-layer ResNet trained on CIFAR-10. Above $90\%$ of sparsity, the value needed to achieve a memory footprint $\leq$40\si{\kilo\byte}, the accuracy curve shows a sudden drop.

This poses a new challenge: \textit{Is there a way to achieve higher compression rates with lower sparsity to preserve accuracy?} Yes indeed, and this work aims at providing a first viable optimization strategy named \textit{EAST} (\underline{E}ncoding-\underline{A}ware \underline{S}parse \underline{T}raining). The rationale is simple, yet effective: find groups of adjacent weights to be pruned rather than pruning single connections. EAST implements a sparse training procedure based on \textit{adaptive group pruning} where the group size \textit{adapts} to the memory constraint minimizing the amount of sparsity needed. 
EAST operates the \textit{LZ4}~\cite{lz4} encoding scheme, which $(i)$ can be implemented with a lightweight routine of few bytes of memory and $(ii)$ guarantees fast decompression and negligible impact on the inference latency; other encoding schemes can be seamlessly applied. The validation is conducted on a state-of-the-art 9-layer ResNet trained on the CIFAR-10 dataset and ported on an Arm Cortex-M4 MCU. A comparison against the same network compressed with a classical weight pruning and encoded with the same LZ4 scheme proves EAST can achieve higher accuracy (up to 8.73\%) when the available memory is very limited (12\si{\kilo\byte} of flash).

\section{Related Works}\label{sec:back}
Weight pruning is a common technique to generate sparse ConvNets. During training, less important connections are gradually removed until a target level of compression is reached. The weight magnitude is the most popular proxy to drive weight selection, with the intuition that collapsing low-value weights to zero affects the prediction accuracy less. Pioneering works in this field~\cite{han2015learning} demonstrated that most of the parameters can be removed (up to 90\%), still keeping the accuracy of the dense model.
Combined with aggressive quantization (down to 2 bits) and weight encoding, sparse ConvNets may achieve impressive compression ratios, up to 49$\times$ depending on the network topology~\cite{han2016deep}. Also, other recent works investigated the bond between weight pruning and quantization~\cite{tung2018clip, grimaldi2019layer, grimaldi2019optimality}, suggesting a joint optimization to determine the best combination of pruning rate and bit-width. While those kinds of strategies are effective with arithmetic units supporting arbitrary bit-widths, general-purpose cores (the focus of this work) provide a limited instruction-set which restricts the choice of the bit-width to few options, i.e., 8-bit for the Cortex-M core used in this work. 

The authors of~\cite{zhu2017prune} showed that \textit{large-sparse} models outperforms \textit{small-dense} models. However, above a certain threshold of sparsity ($>$90\%), the ConvNet may experience dramatic accuracy degradation. This work addresses this issue with an encoding-aware pruning that meets the same memory constraint with fewer weights pruned. 

An interesting study conducted in~\cite{frankle2019lottery} empirically demonstrated that a ConvNet contains an iso-accuracy sub-network that can be discovered via sparse training. This sub-network may represent the smallest achievable implementation that guarantees the highest accuracy, therefore it could be the best starting point for any encoding scheme. However, how to find it under extremely high levels of sparsity still remains an open issue. This work explores a complementary strategy, namely, to maximize the efficiency of the encoding scheme forcing a certain degree of proximity for zeroed weights.

Also, other works experimented pruning at higher-granularity, namely groups of adjacent weights. An example is~\cite{yu2017scalpel}, in which the group size is fixed to match the parallelism of single-instruction multiple-data (SIMD) units to reduce the inference latency. Rather than improving performance, in this work, we aim at decreasing the memory footprint. Specifically, we adopt group pruning to accelerate the compression rate. As a distinctive feature, both the group size and the physical place of the pruned groups adapt during training according to the target memory. Moreover, our strategy is effective also for cores without a SIMD unit (e.g. Cortex-M0 and M3).

Sparse networks can be compressed using different encoding techniques, like \textit{Huffman Coding}~\cite{han2016deep}, \textit{Compressed Sparse Column}~\cite{han2016eie}, \textit{Zero Run Length}~\cite{parashar2017scnn}, to name the most known. In this work we focus on the \textit{LZ4} algorithm which proved faster than others during decompression (in the order of several GB/s on high-end CPUs). For LZ4 to be effective, and in general for any compression algorithm, to have big chunks of adjacent \textit{zeros} is paramount. That is why an efficient compression strategy should take the sparsity distribution (and not just its absolute value) as a first order variable.
\section{Methodology}
\begin{figure}
  \centering
  \subfigure[weight pruning]{%
    \includegraphics[width=0.3\textwidth]{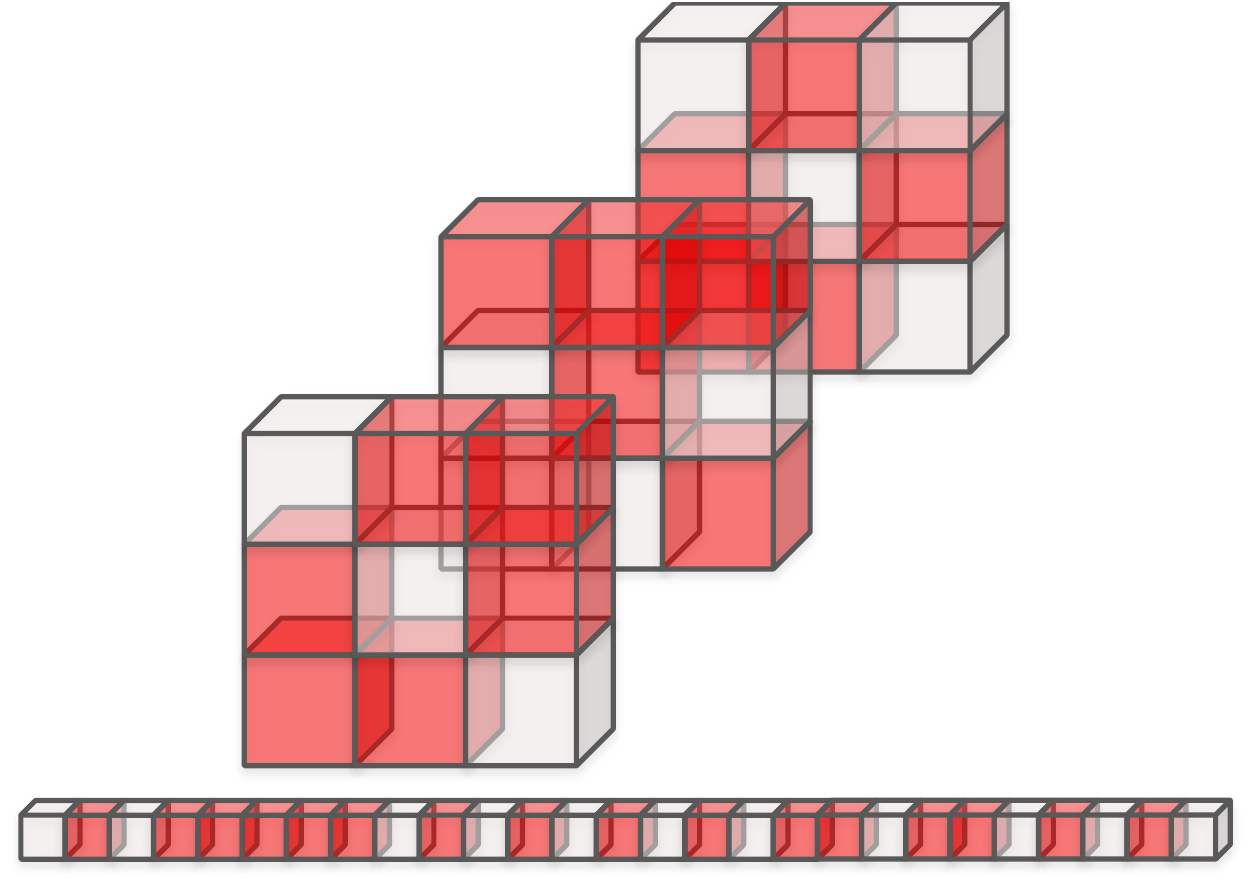}%
    \label{fig:wprune}%
    }\hspace{0.2cm}
    \subfigure[group pruning ($GS=4$)]{%
    \includegraphics[width=0.3\textwidth]{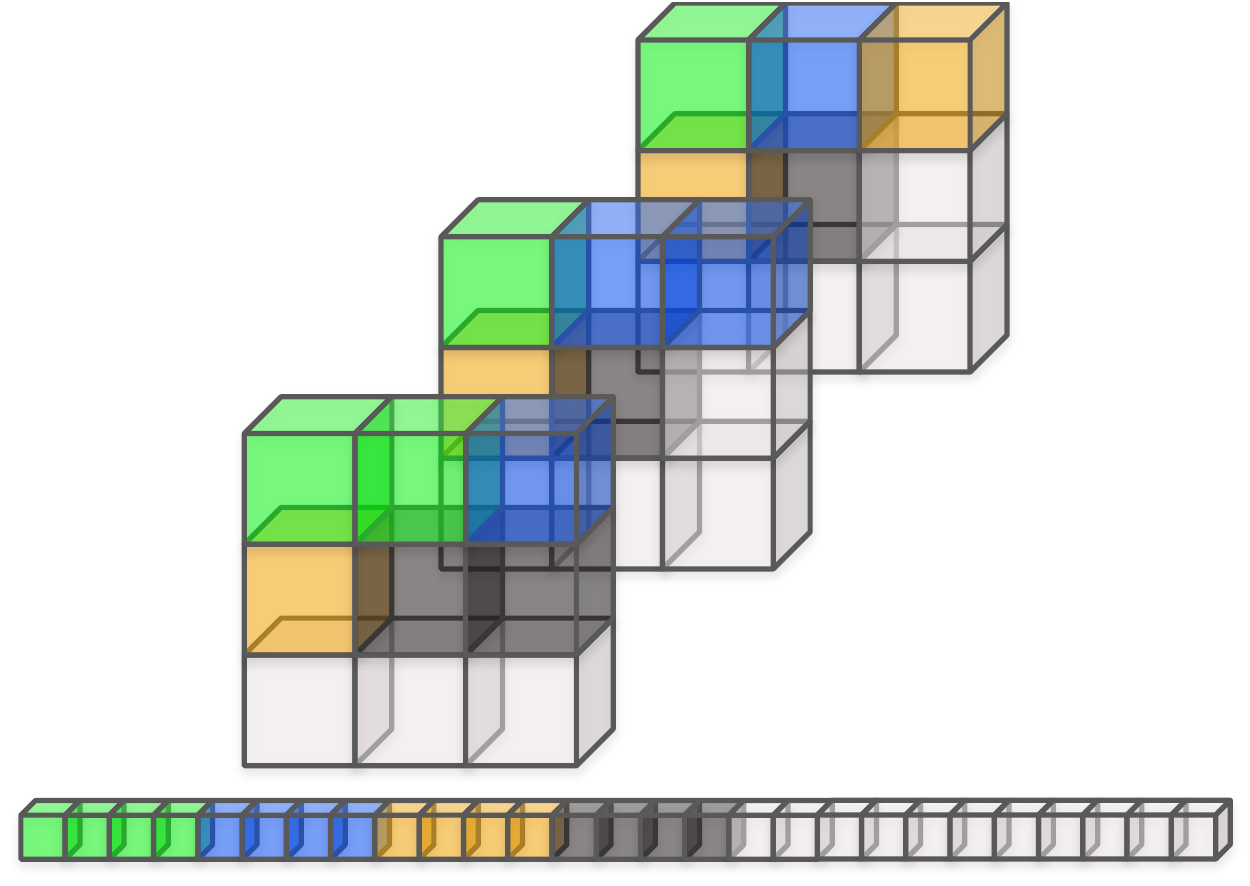}%
    \label{fig:gprune}%
  }%
  \caption{Weight Pruning (a) vs Group Pruning (b). Colored weights denotes zero-values} 
  \label{fig:ab}
  \vspace*{-6mm}
\end{figure}

\subsection{Flow Overview}
The optimization flow encompasses three stages: \textit{(i)} \textit{sparse training}, \textit{(ii)} \textit{quantization}, and \textit{(iii)} \textit{encoding}. The first, \textit{Sparse training}, is to train the sparse network under a user-defined memory constraint. Then, \textit{Quantization} reduces the arithmetic precision of the model to a given bit-width (8-bit in this work). Finally, \textit{Encoding} compresses the model size leveraging favorable patterns made available through the sparse training. The EAST strategy implements the first stage.

\subsection{EAST}\label{sec:method}
\textbf{Encoding-Aware Pruning.} As already introduced in Section~\ref{sec:intro}, accuracy-driven weight pruning algorithms return tensors with sequences of zeros much longer than in the original dense model. Although this helps to increase the compression rate of the encoding scheme, there is no direct control on the position of the zeros, which is ruled by accuracy. The EAST technique is based on the assumption that a weight pruning which is aware of the encoding scheme could make better use of sparsity. A pictorial view of this principle is given in Fig.~\ref{fig:ab}, which illustrates how multi-dimensional tensors are transformed into a 1-D array that can be processed by standard general-purpose cores. Fig. (a) shows a standard weight pruning, while Fig. (b) is for a group pruning with group size ($GS$) of 4. In both cases, the picture refers to a \textit{channel-last} layout organization (the same scheme used by the inference library adopted in this work). The colored items represent the pruned weights. While for the standard method the pruned weights are often placed far away as the selection is just accuracy-driven (smaller weights pruned first), with a group pruning the proximity of the pruned weights is forced by the size of the group itself. This brings to clear advantages. Indeed, even if both cases show the same sparsity (59\%), group pruning gets 55\% higher compression ratio (using LZ4, see Sec. \ref{sec:encoding}). The savings get larger when considering tensors of higher dimensionality. 

It is thereby intuitive that the group size serves as a control knob to reach the best trade-off between accuracy, sparsity, and compression rate. When the available memory is extremely small, groups of larger size may help to achieve higher compression with lower sparsity, hence preserving accuracy more. With a too-small group size, e.g. 1 as for standard weight pruning, the amount of sparsity needed by the encoding algorithm to meet the memory constraint would be too large, with negative effects on the accuracy. The EAST strategy implements a memory-driven adaptive group sizing during the sparse training procedure.

\textbf{Sparse Training.} 
In EAST, both sparsity and group size are gradually increased during the training loop until the memory constraint is met. In the beginning, the sparsity is low and the group size is set to one, hence EAST behaves like a standard weight pruning. 
If the memory constraint is not satisfied, sparsity and group size are updated following a pre-defined schedule. The sparse training iterates for a new bunch of epochs and if the memory constraint is still not met, sparsity and group size are newly updated. The larger the group size, the faster the memory reduction. Therefore, group pruning helps to converge faster attaining the target memory with a lower sparsity. The group selection is driven by the $\ell_2$-norm: groups with lower norm are removed first. However, they can be restored later during the training steps that follow. Once the target memory is met, the sparsity and group size updates are stopped, the pruned weights are frozen, and the training iterates for the last set of epochs adjusting the remaining weights in order to maximize accuracy.

\textbf{Hyperparamters.} Group pruning is applied at the end of each epoch, namely after a complete iteration over the entire training set. The initial target sparsity is 30\% with an increased step of 1\% every epoch; the step is halved at epochs 20 and 50. The initial group size is set to one; starting from epoch 20, it increases with a step of 1 every 10 epochs.

\subsection{Quantization \& Encoding}\label{sec:encoding}
After the sparse training, the 32-bit floating-point ConvNet is quantized to 8-bit. The effect of the quantization is $(i)$ to reduce the memory footprint ensuring marginal accuracy loss, $(ii)$ to increase the frequency of repeated weights, $(iii)$ to accelerate the inference time. We opted for a binary-point quantization scheme which is fully compliant with the inference library used for on-board deployment (CMSIS-NN~\cite{lai2018cmsis}), therefore tailored for the target MCU (the Cortex-M by ARM).

As the very last stage, the quantized model is compressed. EAST can operate different encoding algorithms, but we found the LZ4 algorithm is a good choice for resource-constrained MCUs due to its lightweight routine that ensures high decoding speed. On-board measurements validated this qualitative analysis. The implemented compression strategy is layer-wise, namely, layers are compressed as separate blocks. This solution allows more efficient management of the available SRAM as it avoids one-shot full model decoding. In fact, layers are processed in sequence during inference, therefore each layer block can be decoded independently and temporarily stored in the SRAM using time-shared buffers.
\section{Experimental results}
\subsection{Benchmarks, Datasets, and Training}
We used as benchmark a 9-layer ResNet~\cite{torchskeleton} (ResNet-9) for image classification on the CIFAR-10 dataset.
ResNet-9 currently holds the first position in the DawnBench Competition~\cite{coleman2019analysis}. In our implementation, we removed 75\% of the filter from each convolutional layer. 
As it is already optimized for fast training and inference, this ConvNet represents a challenging test-case to assess the efficiency of different compression pipelines.
The dataset is split in training (45K images), validation (5K) and test (10K) set. The model with the highest accuracy on the validation set is selected for evaluation. For data augmentation, we applied padding with random crop, random horizontal flip, and cutout. The same setting is used for both dense and sparse training. The training is driven by SGD for 200 epochs with batch-size 128. The learning rate follows a cosine annealing schedule with an initial value of 0.1. All the experiments have been run in Pytorch 1.2.

For what concerns quantization, the fixed-point position is determined by a heuristic that minimizes the mean squared error between the floating-point and the 8-bit values. For the intermediate activations, the statistics have been collected on a sub-set of the validation set (size 100 samples). 

Table~\ref{tab:bench} reports the top-1 accuracy on the test set and the memory size of the network. The reported values refer to a standard training (i.e. EAST off). Results confirm the efficiency of quantization (column \textit{Q8}) that gets $4\times$ memory reduction with negligible accuracy losses (0.09\%) w.r.t. the floating-point ConvNet (column \textit{FP32}). Applying the LZ4 compression to the quantized model does not show significant savings: just a few bytes of memory reduction (column \textit{Q8+LZ4}).

\begin{table}[!h]
    \footnotesize
    \centering
    \caption{Top-1 accuracy on CIFAR-10 and weight memory of the dense ResNet-9 after 32-bit floating-point training (FP32), after quantization (Q8), and after LZ4 compression (Q8+LZ4).}
    \vspace{0.2cm}
    \label{tab:bench}
    \begin{tabular}{llll}
    \toprule
     & \textbf{FP32}     & \textbf{Q8}       & \textbf{Q8+LZ4}     \\ \midrule
    \textbf{Top-1} & 91.10\%  & 91.01\%  & 91.01\% \\
    \textbf{Memory} & \SI{558}{\kilo\byte}   & \SI{140}{\kilo\byte}   & \SI{140}{\kilo\byte}   \\ 
    \bottomrule
    \end{tabular}
\end{table}

\subsection{Results}
\begin{table}[t]
    \rowcolors{2}{gray!25}{white}
    \footnotesize
    \centering
    \caption{Sparsity ($S$) and Top-1 Accuracy ($A$) of weight pruning (WP) and EAST on ResNet-9 at different memory constraint $M_\textrm{t}$ (\si{\kilo\byte}). CR is the compression ratio w.r.t. the floating-point ConvNet.}
    \vspace{0.2cm}
    \label{tab:results}
    \resizebox{\columnwidth}{!}{%
    \begin{tabular}{rrllllr}
    \toprule
    \textbf{$M_\textrm{t}$} & \multicolumn{1}{l}{\textbf{CR}} & \textbf{$S_\textrm{WP}$} & \textbf{$S_\textrm{EAST}$} & \textbf{$A_\textrm{WP}$} & \textbf{$A_\textrm{EAST}$} & \multicolumn{1}{l}{\textbf{$\Delta A$}} \\ 
    \midrule
    112 & $5.0\times$ & 58.5\% & 49.5\% & 89.80\% & 89.46\% & -0.34\%\\
    80 & $7.0\times$ & 76.0\% & 60.5\% & 88.67\% & 88.61\% & -0.06\%\\
    48 & $11.6\times$& 89.5\% & 74.8\% & 87.51\% & 87.44\% & -0.07\%\\
    40 & $14.0\times$& 92.0\% & 79.0\% & 86.80\% & 86.82\% & \textbf{0.02}\%\\
    32 & $17.4\times$& 94.0\% & 83.3\% & 85.30\% & 86.11\% & \textbf{0.81}\%\\
    24 & $23.3\times$& 96.0\% & 87.8\% & 82.33\% & 83.65\% & \textbf{1.32}\%\\
    20 & $27.9\times$& 96.8\% & 90.0\% & 79.63\% & 81.11\% & \textbf{1.48}\%\\
    16 & $34.9\times$& 97.5\% & 91.8\% & 74.16\% & 78.45\% & \textbf{4.29}\%\\
    12 & $46.5\times$& 98.3\% & 94.0\% & 55.59\% & 64.32\% & \textbf{8.73}\%\\
    \bottomrule
    \end{tabular}
    }
    \vspace*{-6mm}
\end{table}

\textbf{EAST opens the deep memory space.} Table~\ref{tab:results} reports the comparison between a standard sparse training via weight pruning (WP) and the proposed flow built upon EAST. The two are compared for different target memories ($M_\textrm{t}$). The WP is trained using the same sparsity schedule of EAST (see Sec.~\ref{sec:method}). For each $M_\textrm{t}$, the table collects the compression ratio (CR) achieved after quantization and encoding, the sparsity reached after training (columns $S_\textrm{WP}$ and $S_\textrm{EAST}$), the top-1 accuracy measured on the test-set ($A_\textrm{WP}$ and $A_\textrm{EAST}$) and the relative accuracy distance ($\Delta A$) between EAST and WP.
As demonstrated by previous works, when the memory constraint is met with low sparsity, weight pruning guarantees marginal accuracy losses. For instance, at $M_t =$ \SI{112}{\kilo\byte} the accuracy loss is only 1.21\% lower than the dense 8-bit ConvNet (89.80\% vs 91.01\%). In this region of memory, EAST reaches similar accuracy levels than weight pruning, 0.34\% lower in the worst case ($M_t =$ \SI{112}{\kilo\byte}). However, in the deep memory space ($M_\textrm{t} \leq$ \SI{40}{\kilo\byte}) weight pruning starts experiencing dramatic accuracy degradation. The reason is that very high sparsity ($>90\%$) is needed to reach the desired memory constraint, therefore the model loses its expressive power as only a few weights remain up. In this region EAST outperforms WP; the encoding-aware pruning enables better control of the sparsity indeed ($S_\textrm{EAST} < S_\textrm{WP}$), preserving the same amount of information within the same amount of memory.
On the extreme corner, $M_\textrm{t} = \SI{12}{\kilo\byte}$, EAST is 8.73\% more accurate than WP due to a lower sparsity (94\% vs 98.3\%). To emphasize the role of EAST, one can consider that with the same amount of sparsity (e.g. 94\%) the model optimized with EAST is 2.7$\times$ smaller (row \SI{12}{\kilo\byte} vs \SI{32}{\kilo\byte}).

\begin{figure}[!t]
     \centering
     \epsfig{file=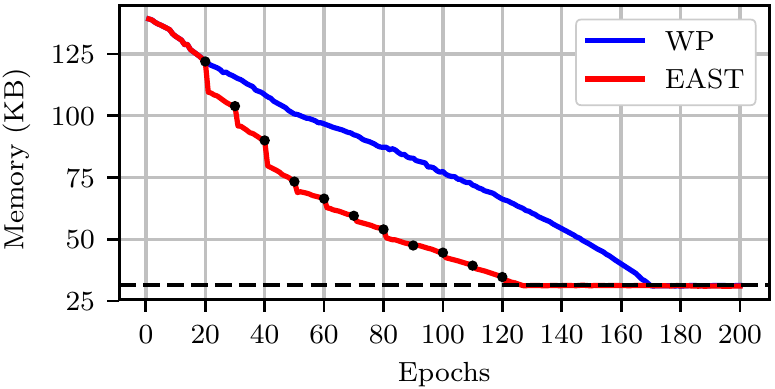,width=0.45\textwidth}
     \caption{Epochs vs. Memory in weight pruning (blue line) and EAST (red line) for $M_\textrm{t} = $ \SI{32}{\kilo\byte} (dashed line). The dots indicates when the group size increases.} 
     \label{fig:memory}
    \vspace*{-6mm}
\end{figure}

\textbf{EAST accelerates the memory compression.} Fig.~\ref{fig:memory} shows the evolution of the memory footprint during the training epochs for both WP (blue line) and EAST (red line) under the same memory constraint $M_\textrm{t} =$ \SI{32}{\kilo\byte}. During the first 20 epochs, when the group size is one (as set by the training schedule, see Sec. \ref{sec:method}), EAST follows the same trend of WP. Every time the group size gets increased (events indicated with black dots), the memory compression accelerates quickly. As a result, EAST reaches the target memory (indicated with the dashed black line) 43 epochs sooner than WP. These findings suggest that the group size works as an effective knob to boost the compression rate without seeking additional sparsity.

\textbf{Efficient deployment of sparse ConvNets.} We validated the optimization flow on a STM32 NUCLEO-F412ZG \cite{nucleo} board powered with an Arm Cortex-M4 core running at \SI{100}{\mega\hertz}, \SI{1}{\mega\byte} of flash memory, and \SI{512}{\kilo\byte} of SRAM. As the inference engine, we adopted the CMSIS-NN library. 
The original dense ConvNet takes \SI{28}{\kilo\byte} of SRAM to store intermediate activations and classifies a single image in \SI{492}{\milli\second}. The sparse ConvNets needs \SI{884}{\byte} of flash for the LZ4 routine, which thereby has a negligible impact on the compression rates achieved. Furthermore, an additional SRAM buffer of \SI{36}{\kilo\byte} is needed to store the decompressed weights. Since this buffer is time-shared among different layers, its size is given by the biggest layer. However, the buffer can be dynamically allocated just before the execution of the ConvNet.

The total execution time is function of the memory constraint $M_\textrm{t}$: the larger the $M_\textrm{t}$, the longer the decompression stage. For ResNet-9 generated with EAST, the execution time ranges from \SI{482}{\milli\second} at $M_\textrm{t} =$ \SI{12}{\kilo\byte} to \SI{497}{\milli\second} at $M_\textrm{t} =$ \SI{112}{\kilo\byte}. At the lowest memory, the decompression only accounts for \SI{6}{\milli\second}; in all cases, the network layers execute faster than the dense counterpart as the weights resides in the SRAM instead of flash.

\subsection{Final Remark}
This work opens new paths towards the optimization of ConvNets in memory-bounded cores. EAST is particularly suited for the deep memory space, where it outperforms state-of-the-art sparse training. Nevertheless, further investigation is needed to bridge the accuracy gap with dense nets at extreme constraints. First, we plan to consider other proxies than the $\ell_2$-norm to drive the group selection. Second, group size and sparsity follow a relative straightforward scheduling during the training; in order to achieve better trade-offs between sparsity, group size, and the position of the pruned groups, future works will explore the adoption of smarter hyper-parameter tuning techniques (e.g. Bayesian optimization or reinforcement learning) that might help EAST to reach global optima in the sparsity-memory-accuracy space. 
\section{Conclusions}
EAST is a novel strategy for the training of memory-bounded sparse ConvNets. Leveraging the working principle upon which the encoding algorithms are built, it trains sparse networks that are more amenable to compression, yet less sparse and thus more accurate.
The collected results are promising; EAST reaches up to $46.5\times$ compression w.r.t. the original floating-point model achieving 8.73\% higher accuracy than state-of-the-art sparse training.  Also, we presented an efficient per-layer compression strategy, which exploits the proposed sparse training on commercial off-the-shelf IoT cores.

\bibliographystyle{IEEEtran}
\bibliography{refs}

\begin{thebibliography}{10}
\providecommand{\url}[1]{#1}
\csname url@samestyle\endcsname
\providecommand{\newblock}{\relax}
\providecommand{\bibinfo}[2]{#2}
\providecommand{\BIBentrySTDinterwordspacing}{\spaceskip=0pt\relax}
\providecommand{\BIBentryALTinterwordstretchfactor}{4}
\providecommand{\BIBentryALTinterwordspacing}{\spaceskip=\fontdimen2\font plus
\BIBentryALTinterwordstretchfactor\fontdimen3\font minus
  \fontdimen4\font\relax}
\providecommand{\BIBforeignlanguage}[2]{{%
\expandafter\ifx\csname l@#1\endcsname\relax
\typeout{** WARNING: IEEEtran.bst: No hyphenation pattern has been}%
\typeout{** loaded for the language `#1'. Using the pattern for}%
\typeout{** the default language instead.}%
\else
\language=\csname l@#1\endcsname
\fi
#2}}
\providecommand{\BIBdecl}{\relax}
\BIBdecl

\bibitem{seidenari2017deep}
L.~Seidenari, C.~Baecchi, T.~Uricchio, A.~Ferracani, M.~Bertini, and A.~D.
  Bimbo, ``Deep artwork detection and retrieval for automatic context-aware
  audio guides,'' \emph{ACM Transactions on Multimedia Computing,
  Communications, and Applications (TOMM)}, vol.~13, no.~3s, p.~35, 2017.

\bibitem{lai2018cmsis}
L.~Lai, N.~Suda, and V.~Chandra, ``Cmsis-nn: Efficient neural network kernels
  for arm cortex-m cpus,'' \emph{arXiv preprint arXiv:1801.06601}, 2018.

\bibitem{han2015learning}
S.~Han, J.~Pool, J.~Tran, and W.~Dally, ``Learning both weights and connections
  for efficient neural network,'' in \emph{Advances in neural information
  processing systems}, 2015, pp. 1135--1143.

\bibitem{zhu2017prune}
M.~Zhu and S.~Gupta, ``To prune, or not to prune: exploring the efficacy of
  pruning for model compression,'' \emph{arXiv preprint arXiv:1710.01878},
  2017.

\bibitem{han2016deep}
S.~Han, H.~Mao, and W.~J. Dally, ``Deep compression: Compressing deep neural
  network with pruning, trained quantization and huffman coding,'' in \emph{4th
  International Conference on Learning Representations, {ICLR} 2016}, 2016.

\bibitem{tung2018clip}
F.~Tung and G.~Mori, ``Clip-q: Deep network compression learning by in-parallel
  pruning-quantization,'' in \emph{Proceedings of the IEEE Conference on
  Computer Vision and Pattern Recognition}, 2018, pp. 7873--7882.

\bibitem{lz4}
\BIBentryALTinterwordspacing
Lz4. [Online]. Available: \url{https://lz4.github.io/lz4/}
\BIBentrySTDinterwordspacing

\bibitem{grimaldi2019layer}
M.~Grimaldi, V.~Tenace, and A.~Calimera, ``Layer-wise compressive training for
  convolutional neural networks,'' \emph{Future Internet}, vol.~11, no.~1,
  p.~7, 2019.

\bibitem{grimaldi2019optimality}
M.~{Grimaldi}, V.~{Peluso}, and A.~{Calimera}, ``Optimality assessment of
  memory-bounded convnets deployed on resource-constrained risc cores,''
  \emph{IEEE Access}, vol.~7, pp. 152\,599--152\,611, 2019.

\bibitem{frankle2019lottery}
J.~Frankle, G.~K. Dziugaite, D.~M. Roy, and M.~Carbin, ``The lottery ticket
  hypothesis at scale,'' \emph{arXiv preprint arXiv:1903.01611}, 2019.

\bibitem{yu2017scalpel}
J.~Yu, A.~Lukefahr, D.~Palframan, G.~Dasika, R.~Das, and S.~Mahlke, ``Scalpel:
  Customizing dnn pruning to the underlying hardware parallelism,'' in
  \emph{ACM SIGARCH Computer Architecture News}, vol.~45, no.~2.\hskip 1em plus
  0.5em minus 0.4em\relax ACM, 2017, pp. 548--560.

\bibitem{han2016eie}
S.~Han, X.~Liu, H.~Mao, J.~Pu, A.~Pedram, M.~A. Horowitz, and W.~J. Dally,
  ``Eie: efficient inference engine on compressed deep neural network,'' in
  \emph{2016 ACM/IEEE 43rd Annual International Symposium on Computer
  Architecture (ISCA)}.\hskip 1em plus 0.5em minus 0.4em\relax IEEE, 2016, pp.
  243--254.

\bibitem{parashar2017scnn}
A.~Parashar, M.~Rhu, A.~Mukkara, A.~Puglielli, R.~Venkatesan, B.~Khailany,
  J.~Emer, S.~W. Keckler, and W.~J. Dally, ``Scnn: An accelerator for
  compressed-sparse convolutional neural networks,'' in \emph{2017 ACM/IEEE
  44th Annual International Symposium on Computer Architecture (ISCA)}.\hskip
  1em plus 0.5em minus 0.4em\relax IEEE, 2017, pp. 27--40.

\bibitem{torchskeleton}
\BIBentryALTinterwordspacing
Torchskeleton. [Online]. Available:
  \url{https://github.com/wbaek/torchskeleton}
\BIBentrySTDinterwordspacing

\bibitem{coleman2019analysis}
C.~Coleman, D.~Kang, D.~Narayanan, L.~Nardi, T.~Zhao, J.~Zhang, P.~Bailis,
  K.~Olukotun, C.~R{\'e}, and M.~Zaharia, ``Analysis of dawnbench, a
  time-to-accuracy machine learning performance benchmark,'' \emph{ACM SIGOPS
  Operating Systems Review}, vol.~53, no.~1, pp. 14--25, 2019.

\bibitem{nucleo}
\BIBentryALTinterwordspacing
Nucleo-f412zg. [Online]. Available:
  \url{https://www.st.com/en/evaluation-tools/nucleo-f412zg.html}
\BIBentrySTDinterwordspacing

\end{thebibliography}

\end{document}